# A Visual Information Extraction System Using Multimodal Large Language Models for Scanned Cognitive and Behavioral Assessments

**Enshuo Hsu[1, 2], Linh Nguyen[3], PhD, Sunyang Fu[4], PhD, Heath Goodrum[3], Sandra Stansberry[3], Kirk Roberts[1], PhD, Hongfang Liu[4], PhD & Yen-Chi L. Le[3], PhD**

[1]McWilliams School of Biomedical Informatics, The University of Texas Health Science Center at Houston, Houston, Texas, USA

[2]Enterprise Development and Integration, University of Texas MD Anderson Cancer Center, Houston, Texas, USA

[3]UT Physicians Center for Population Health Management & Quality, The University of Texas Health Science Center at Houston, Houston, Texas, USA

[4]Department of Quantitative and Systems Health Sciences, Dell Medical School, The University of Texas at Austin, Austin, TX, USA.

**Corresponding author:** Yen-Chi Le, PhD, UTHealth Houston | UT Physicians, 6410 Fannin Suite 810, Houston, TX 77030

**Keywords:**

Large language models, Optical character recognition, Natural language processing, Electronic health records

**Word Count:**

Total: 3967/4000

Abstract: 245/250

## ABSTRACT

**Objective**. Paper-based assessments are common in patient care, but handwriting and scanning artifacts make these forms difficult to process at scale. This study proposes a modular, multimodal LLM-based system with human oversight for extracting fields and scores. We conducted a qualitative analysis of the use of LLMs for hand-drawing analysis. **Materials and Methods**. The system processes cognitive assessments (i.e., MMSE, MoCA, Mini-Cog) and the NICHQ Vanderbilt Assessment using two modules: 1) a form classifier that assigns a scanned page to a form type, followed by a corresponding 2) score extractor that extracts fields and scores into a schema. A human-in-the-loop design utilizing dual-model validation engages human annotators when two LLMs disagree, limiting manual review to challenging pages only. **Results**. End-to-end accuracy for cognitive assessments was 88%, with most form types exceeding 90%. For the NICHQ Vanderbilt, overall accuracy reached 97.5%, with all form types above 96%. The human-in-the-loop design further improved cognitive assessment accuracy to 97.5% while reducing manual work by 63.8%. In the Mini-Cog Clock Drawing task, qualitative analysis showed that 62.3% of all extraction errors stemmed from ambiguous drawing, in which cases the LLM provided reasonable readings. **Discussion**. Compared to current manual data extraction, our LLM-based system reduces processing time by 10 to 20 times and costs by over 12 times. The modular design allows seamless expansion to additional form types. **Conclusion**. Our system provides an accurate, scalable, and cost-efficient solution for structuring data trapped in handwritten paper forms for clinical decision support.

## OBJECTIVE

Paper-based assessment forms (e.g., instruments) with handwritten text and marks are widely used in general clinical practice for collecting patient information and are heavily adopted in medical specialties such as neurology, psychiatry, and geriatrics. Like other types of scanned medical documents, these forms are hard to process due to scanning artifacts (e.g., rotation, skewness, blur, and background noise). In addition, their handwritten nature and inconsistent layout introduce extended challenges [1–3]. In current practice, retrieving information often requires manual review of individual scans, which is not only labor-intensive and time-consuming but also unscalable, given the large volume in electronic health records (EHRs) of major hospital systems [3]. In 2024, the Agency for Healthcare Research and Quality (AHRQ) reaffirmed that unstructured EHR data, including scanned documents, clinical notes, and difficult-to-interpret reports, pose substantial risks to diagnostic safety and continuity of care [4]. An automatic information extraction system for paper-based forms is urgently needed.

Previous studies adopted traditional Computer Vision (CV)-based methods for Optical Character Recognition (OCR) and Optical Mark Recognition (OMR) in scanned medical document processing and reported challenges with handwriting, tabular layouts, and low-quality scans [2,5]. Recently, as multimodal Large Language Models (LLMs) show dominant performance on general-domain OCR tasks, a few studies have adopted them for visual information extraction from scanned transfusion reaction reports [6] and death certificates [7]. However, multimodal LLM adoption is in its early phase, with its generalizability yet to be evaluated. Furthermore, existing studies are mostly evaluated on a single document type with a known layout. The robustness of multimodal LLMs on varying types of forms has not been thoroughly evaluated. Finally, current studies on information extraction of cognitive assessments have mostly focused on text-based retrieval of manually transcribed scores from clinical notes [8–10]. To our knowledge, automatic extraction from the scanned sources has not been proposed. Therefore, this study proposed a multimodal LLM-based system for end-to-end processing of paper-based forms. We evaluated the system performance, error patterns, processing time, and costs using the most widely adopted cognitive screening instruments (i.e., Mini-Mental State Examination (MMSE), the Montreal Cognitive Assessment (MoCA), and the Mini-Cog) and the Attention-deficit/hyperactivity disorder (ADHD) instrument (i.e., National Institute for Children's Health Quality Vanderbilt ADHD Rating Scale).

This study has the following significance: 1) As a pioneering work, it evaluates the use of multimodal LLMs for visual information from handwritten and hand-marked forms, commonly used in the current healthcare practice. 2) It qualitatively evaluates the use of LLMs for hand-drawing analysis, using the Mini-Cog clock-drawing task as an example. 3) It proposes an automatic system, comprising a form classifier and score extractors, for an end-to-end structured data extraction from widely used cognitive and ADHD assessments. 4) It proposes a practical human-in-the-loop design with dual-model validation that reduces manual work by two-thirds while achieving high accuracy for cognitive assessments.

# BACKGROUND AND SIGNIFICANCE

## Visual Information Extraction with Multimodal Large Language Models

Multimodal LLMs have been adopted for OCR tasks and have shown remarkable performance on general-domain OCR benchmarks [11,12]. In the biomedical field, Schäfer et al. proposed an OCR pipeline comprising a YOLO-based checkbox detection module followed by a Vision Language Model-based holistic matching module for checked text category detection from scanned transfusion reaction reports [6]. Our recent work evaluated two multimodal LLM-based methods for cause-of-death ascertainment from scanned death certificates [7]. Despite growing interest, multimodal LLM adoption in the biomedical field remains in its early phase. To our knowledge, its use in clinical scanned-document processing, including OCR and OMR, has not been thoroughly evaluated.

## Automatic Information Extraction for Psychometric Instruments

In the clinical natural language processing (NLP) field, a few studies adopted text-based NLP methods to systematically extract psychometric instrument fields and scores. Wang et al. evaluated a lexicon and terminology-based NLP system to extract Mini-Mental State Examination (MMSE) and Montreal Cognitive Assessment (MoCA) scores from clinical notes within an EHR [8]. Ouyang et al. developed a rule-based NLP method to extract Saint Louis University Mental Status (SLUMS) Examination scores from clinical notes [9]. Recently, Zhang et al. utilized ChatGPT and Llama 2 for cognitive exam score extraction, including MMSE, Clinical Dementia Rating (CDR), and MoCA, from clinical notes [10].

Most existing studies focused on extracting the transcribed, text-form psychometric instrument scores from clinical notes. To our knowledge, automatic score extraction from source handwritten instruments has not been thoroughly evaluated.

### Assessments for Cognitive Function

Cognitive assessments are standard tools evaluating memory, attention, language, and orientation to support the diagnosis and longitudinal monitoring of cognitive impairment. Three widely used cognitive screening instruments are the Mini-Mental State Examination (MMSE) [13], the Montreal Cognitive Assessment (MoCA) [14], and the Mini-Cog [15]. The MMSE and MoCA provide global measures of cognitive function, while the Mini-Cog is a brief screening instrument. See Figure 1 (A)-(D) for examples of these scanned assessments.

### Assessment for the Diagnosis and Management of Attention Deficit Hyperactivity Disorder (ADHD)

The National Institute for Children's Health Quality (NICHQ) Vanderbilt ADHD Rating Scale (V-ADHD-RS) is a widely used tool for the assessment and management of ADHD in pediatrics [16,17]. It is completed by parents and teachers evaluating the frequency and severity of inattentive and hyperactive-impulsive behaviors using a four-point Likert scale ranging from 0 ("never") to 3 ("very often") across several domains. Respondents also provide contextual information (e.g., free-text comments). See Figure 1 (E)-(H) for examples of scanned Vanderbilt Assessment.

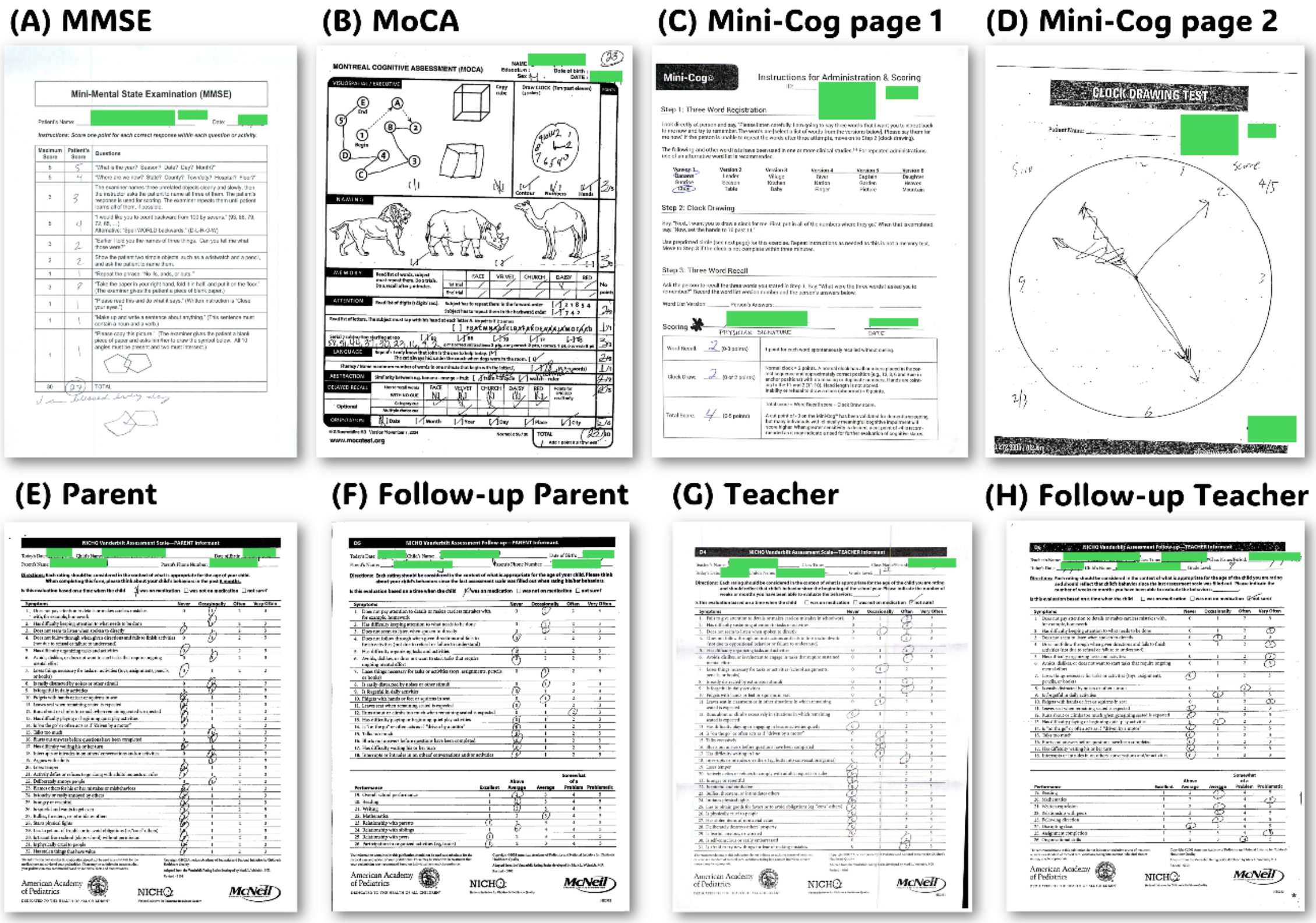

***Figure 1: Scanned cognitive assessments and NICHQ Vanderbilt Assessment.*** *(A) The Mini-Mental State Examination (MMSE). (B) The Montreal Cognitive Assessment (MoCA). (C) The Mini-Cog instruction and scoring page. (D) The Mini-Cog clock-drawing page. (E) The NICHQ Vanderbilt Assessment Scale – Parent Information. (F) The NICHQ Vanderbilt Assessment Follow-up – Parent Information. (G) The NICHQ Vanderbilt Assessment Scale – Teacher Information. (H) The NICHQ Vanderbilt Assessment Follow-up – Teacher Information.*

# METHODS

## Methodology Overview

This study developed a modular design for visual information extraction systems to process cognitive assessments (i.e., MMSE, MoCA, Mini-Cog) and the NICHQ Vanderbilt Assessment using multimodal LLMs. The systems begin with form-type classification, which assigns a scanned page to a form type (e.g., MMSE, MoCA, Mini-Cog, or other). The page is then assigned to the corresponding score extraction module for question, field, and score extraction (e.g., Mini-Cog word recall, clock drawing, and total scores). This section describes the LLM configuration, system design, and evaluation (Figure 2). More details are available in the Supplementary *Implementation Details* section.

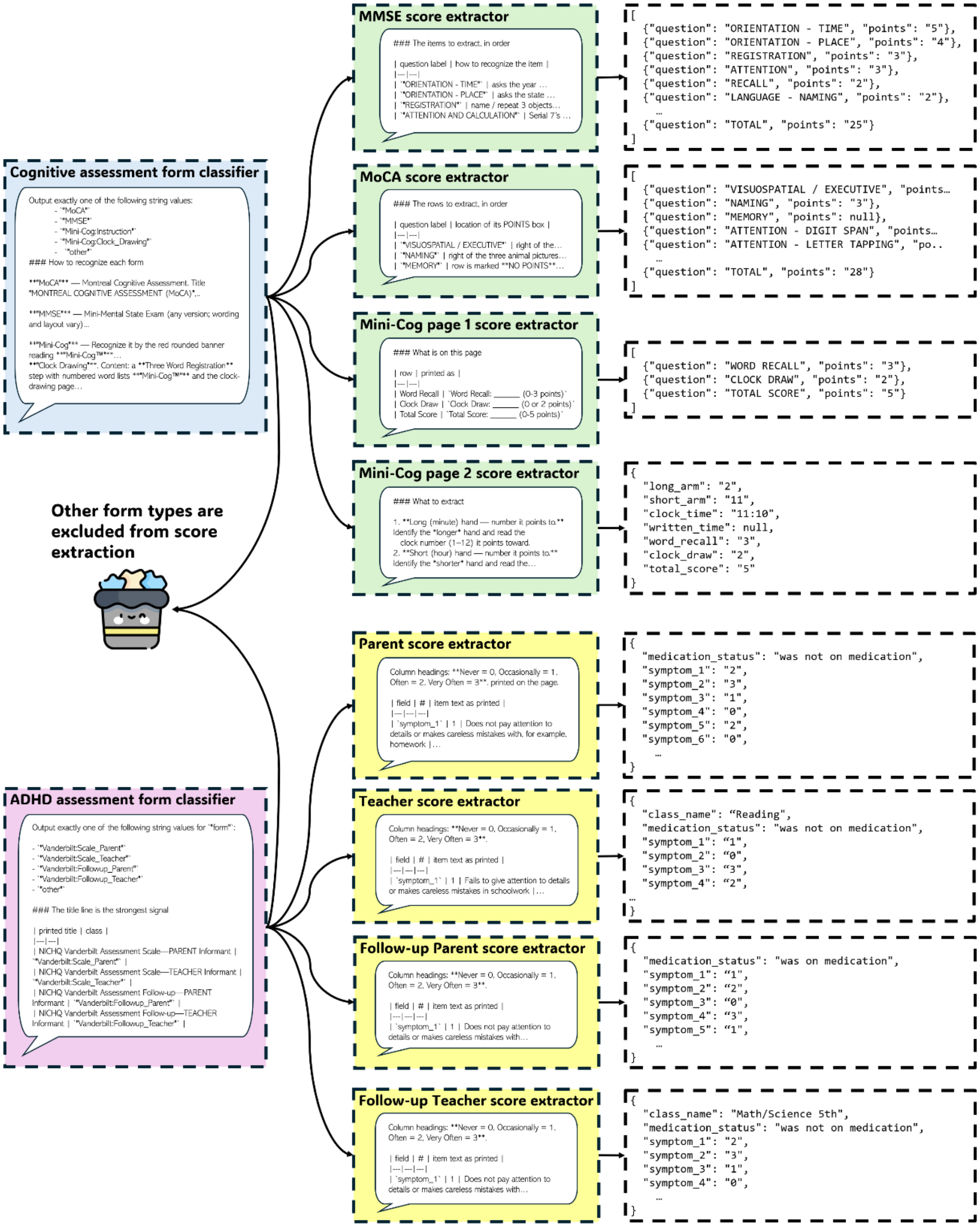

***Figure 2: Methodology flowchart.*** *The Cognitive Assessment Score Extraction System and the NICHQ Vanderbilt Assessment Score Extraction System comprise two layers: form classifier and score extractors. The form classifier assigns a scanned page to a downstream score extractor for question, field, and score extraction.*

## Multimodal Large Language Models

The LLMs, GPT-4.1-mini and GPT-5.4-mini, are accessed via the University of Texas Health Science Center (UTHealth) Azure OpenAI deployments. All data processing and transmission comply with the Health Insurance Portability and Accountability Act (HIPAA). No de-identification or removal of protected health information (PHI) was required. Details are described in the Supplementary *Implementation Details — Multimodal Large Language Models* section.

## Visual Information Extraction System Design

This section provides an overview of our visual information extraction system design. Details are described in the Supplementary I*mplementation Details — Visual Information Extraction System Design* section. We proposed a modular information extraction system comprising a form classifier module and downstream score-extractor modules. **Form classifiers** prompt an LLM to classify a scanned page as one of: "MMSE", "MoCA", "Mini-Cog: Instruction", "Mini-Cog: Clock Drawing", or "Other" for cognitive assessments and "Parent", "Teacher", "Follow-up Parent", "Follow-up Teacher", or "Other" for NICHQ Vanderbilt. Note that the cognitive assessments and NICHQ Vanderbilt Assessment are processed by separate systems. After classification, scanned pages are assigned to the corresponding score extractor. **Score extractors** prompt LLMs to extract the questions, fields, and written scores in a scanned page. They are engineered for respective form types. For example, the MMSE score extractor, the 11 tasks (e.g., Orientation, Attention and Calculation, and Registration), and score ranges are detailed in the prompt template as a Markdown table. The complete prompt templates and system prompts are presented in the Supplementary Tables S13 and S14.

In addition to zero-shot prompting, we evaluated one-shot prompting to further improve system performance. For cognitive assessments, the form classifier was provided with an example Mini-Cog Clock Drawing page with ground-truth labels, randomly selected from the development set. The reason is that we observed worse classification performance on that form type under the zero-shot setting. For each score extractor, a one-shot example of the corresponding type was appended to the prompts. For the NICHQ Vanderbilt Assessment information extraction system, given its near-perfect classification performance, we did not perform one-shot prompting for the classifier.

## Data Source

We extracted 6,699 scanned document files from the EHRs of UT Physician, with document types of ADHD Assessment (n=3,030) and Cognitive Assessment (n=3,669) from 6,578 unique patients from June 1, 2007, to November 4, 2025. The study was reviewed by the University of Texas Health Science Center at Houston Committee for the Protection of Human Subjects (CPHS) and determined to be exempt from human subjects research oversight.

## Cohort Selection and Human Annotation

To create a ground truth for model development and evaluation, three research assistants manually annotated 1000 pages with Label Studio version 1.23.0 [18]. The workflow was comprised of three phases: 1) **form inclusion**, in which the annotators sequentially reviewed and assigned pages to the corresponding form types until at least 100 pages were available for each type, 2) **inter-annotator agreement assessment**, in which the 10 pages of each type were drawn for all three annotators to annotate independently and the Fleiss Kappa score was calculated, and 3) **independent score annotation**, in which each annotator labeled the fields and scores of 30 pages of each type, resulting in a total of 100 pages per type (development set = 5, test set = 95) (Figure 3). More details are available in the Supplementary *Implementation Details — Cohort Selection and Human Annotation* section.

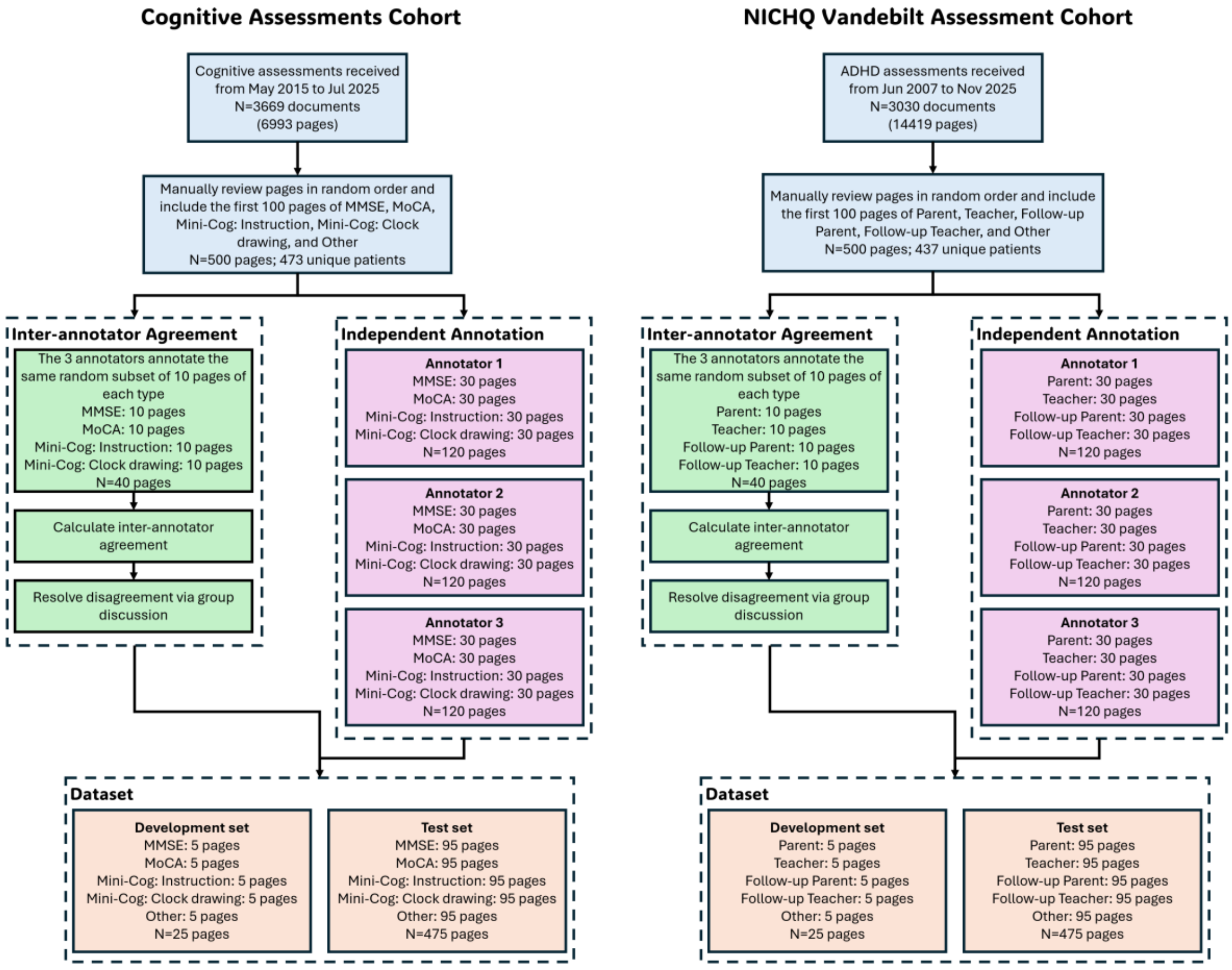

***Figure 3: Manual annotation workflow**. A total of 6699 scanned assessments were collected. Three annotators reviewed and included the first 100 pages of each form type, resulting in cohorts of 1000 pages. Inter-annotator agreements were assessed on a 10% subset of each type. The final datasets for both cognitive assessments and NICHQ Vandebilt Assessment consist of development sets of 25 pages and test sets of 475 pages.*

### Evaluation Metrics

**Mean per-page field accuracy** is used as the primary metric, defined as:

$$Acc = \frac{\#\ of\ correct\ scores\ in\ a\ page}{\#\ of\ total\ scores\ in\ a\ page}$$

For example, an MMSE page has 12 scores in total: 11 itemized scores and 1 total score. A model that correctly extracted 10 of them will be rated,

$$Acc = \frac{10}{12} = 0.833$$

For form classification, precision, recall, and F1 are used. For text-field extraction (i.e., comments and class name in NICHQ Vanderbilt), since they are not always present, we evaluate 1) the presence extraction with precision and recall, and 2) the text extraction with normalized Levenshtein similarity. Details are described in the Supplementary *Implementation – Evaluation Metrics* section.

### Experimental Design

To assess the system's realistic performance while obtaining insights on the impact of each module, we performed two series of experiments: 1) **End-to-end experiments** evaluate the system's overall performance. On occasions that a form is misclassified, a page-level accuracy of zero is automatically assigned, disregarding the score extraction results. In this setting, both the classification and score extraction steps affect the final accuracy and 2) **Modular experiments** evaluate the classification modules and score-extractor modules separately. Documents are sent to the form classifier for classification evaluation. In a separate pipeline, documents are sent to the corresponding score extractors for score extraction. In this setting, the classification step has no impact on the score extraction. The results reflect the performance of each module alone.

**Baseline**. For comparison, a baseline method using the traditional OCR engine Tesseract 4.1.1 [19] was evaluated. The scanned pages were first processed by Tesseract to generate layout-preserving OCR text. The OCR text, rather than the scanned image, was then fed into GPT-5.4-mini for text-based classification and score extraction, using the same schema and similar prompts as our proposed system. The prompts are available in Supplementary Tables S15 and S16.

### Error Analysis

We performed error analysis on the results of GPT-5.4-mini (one-shot) for both the cognitive assessment information extraction system and NICHQ Vanderbilt Assessment information extraction system. Each erroneous field was manually reviewed and assigned a cause based on human judgment (e.g., overwritten or erased mark, model hallucinated a value).

### Human-in-the-loop Design for Cognitive Assessments

To maximize production performance for cognitive assessment score extraction while reducing annotation labor, we propose a human-in-the-loop design that utilizes LLMs to process and validate most scans, allowing human annotators to focus on cases where the models disagreed. The design is comprised of two steps: 1) **Dual-model agreement,** Two LLMs (i.e., GPT-4.1-mini and GPT-5.4-mini) work independently to process the scans end-to-end. When a page-level agreement was achieved (i.e., both models completely agree on the type and scores of a page), the result is automatically accepted. 2) **Human verification**. When the two LLMs disagree, human annotators are engaged "in the loop" to make the final decisions. To evaluate this design, we calculated the 1) proportion of dual-model agreement ("coverage"), which reflects the percentage of labor avoided. 2) Accuracy on agreed pages, which reflects the performance of the validated results by the two models.

### Cost Analysis

We estimated manual annotation time and cost for a hypothetical cohort of 10,000 cognitive assessment forms using observed annotation times and national wage estimates for Medical Transcriptionists from the U.S. Bureau of Labor Statistics, May 2025 Occupational Employment and Wage Statistics [20]. The Supplementary *Implementation Details — Cost Analysis* section provides detailed assumptions, cost calculations, and sensitivity analyses.

### Computation

All experiments were conducted with Python 3.12.12. Visual information extraction systems were implemented with the VLM4OCR Python package version 0.6.0 [21]. The baseline, text-based information extraction was implemented with LLM-IE version 1.4.3 [22].

## RESULTS

### Annotation Overview

The three annotators manually labeled the 10 shared pages for each type to assess inter-annotator agreement. The Fleiss Kappa scores range from 0.8119 to 1.000 for cognitive assessments and from 0.8101 to 1.000 for the NICHQ Vanderbilt assessment, indicating almost perfect agreement. The median total scores are 24, 24, and 5 for MMSE, MoCA, and Mini-Cog, respectively (Table 1). The full list of summary statistics is reported in the Supplementary Tables S1-S8.

*Table 1: Annotation overview*

| | Overall | Development set | Test set |
|---|---|---|---|
| **Cognitive Assessments** | | | |
| N pages | 500 | 25 | 475 |
| N patients | 473 | 25 | 451 |
| **Mini-Mental State Examination (MMSE)** | | | |
| N pages | 100 | 5 | 95 |
| N patients | 100 | 5 | 95 |
| Total score, Median [Q1-Q3] | 24 [18-28] | 22 [15-23] | 25 [18.75-28] |
| **Montreal Cognitive Assessment (MoCA)** | | | |
| N pages | 100 | 5 | 95 |
| N patients | 100 | 5 | 95 |
| Total score, Median [Q1-Q3] | 24 [19-27] | 20 [17-26] | 24 [19-27] |
| **Mini-Cog instruction/scoring page (1st page)** | | | |
| N pages | 100 | 5 | 95 |
| N patients | 100 | 5 | 95 |
| Total score, Median [Q1-Q3] | 5 [4-5] | 5 [5-5] | 5 [4-5] |
| **Mini-Cog clock-drawing page (2nd page)** | | | |
| N pages | 100 | 5 | 95 |
| N patients | 100 | 5 | 95 |
| Total score, Median [Q1-Q3] | 5 [5-5] | 5 [5-5] | 5 [4.75-5] |
| **Other** | | | |
| N pages | 100 | 5 | 95 |
| N patients | 91 | 5 | 86 |
| **NICHQ Vanderbilt Assessment** | | | |
| N pages | 500 | 25 | 475 |
| N patients | 437 | 25 | 417 |
| **NICHQ Vanderbilt Assessment Scale – Parent Information** | | | |
| N pages | 100 | 5 | 95 |
| N patients | 96 | 5 | 92 |
| **NICHQ Vanderbilt Assessment Scale – Teacher Information** | | | |
| N pages | 100 | 5 | 95 |
| N patients | 94 | 5 | 90 |
| **NICHQ Vanderbilt Assessment Follow-up – Parent Information** | | | |
| N pages | 100 | 5 | 95 |
| N patients | 86 | 5 | 82 |
| **NICHQ Vanderbilt Assessment Follow-up – Teacher Information** | | | |
| N pages | 100 | 5 | 95 |
| N patients | 89 | 5 | 85 |
| **Other** | | | |
| N pages | 100 | 5 | 95 |
| N patients | 91 | 5 | 86 |

**End-to-end Score Extraction**

In Cognitive assessments, the GPT-5.4-mini with one-shot achieved the best end-to-end score extraction accuracy of 0.975, 0.911, 0.947, and 0.687 for MMSE, MoCA, Mini-Cog Instruction page, and Mini-Cog clock-drawing page, respectively. The GPT-5.4-mini achieved superior performance to GPT-4.1-mini across most assessments, except on the Mini-Cog clock-drawing page. The one-shot prompting method consistently improves accuracy across LLMs and assessments by 4.1%, 6.8%, 9.8%, and 7.5% for GPT-4.1-mini and 0.5%, 1.8%, 1.0%, and 23.1% for GPT-5.4-mini. Both LLMs dramatically outperformed the Tesseract baseline. In the NICHQ Vanderbilt assessment, the GPT-5.4-mini with zero-shot achieved the best end-to-end score extraction accuracy of 0.975, 0.982, 0.981, and 0.964 for the Parent, Teacher, Follow-up Parent, and Follow-up Teacher forms, respectively. The GPT-5.4-mini zero-shot outperformed the GPT-4.1-min zero-shot by 17.9%, 21%, 16.8%, and 13.8%. One-shot prompting improved performance on GPT-4.1-mini by 10.6%, 11.5%, 8.3%, and 5.3%. However, with GPT-5.4-mini, one-shot prompting yielded similar performance. Both LLMs outperformed the Tesseract baseline (Figure 4).

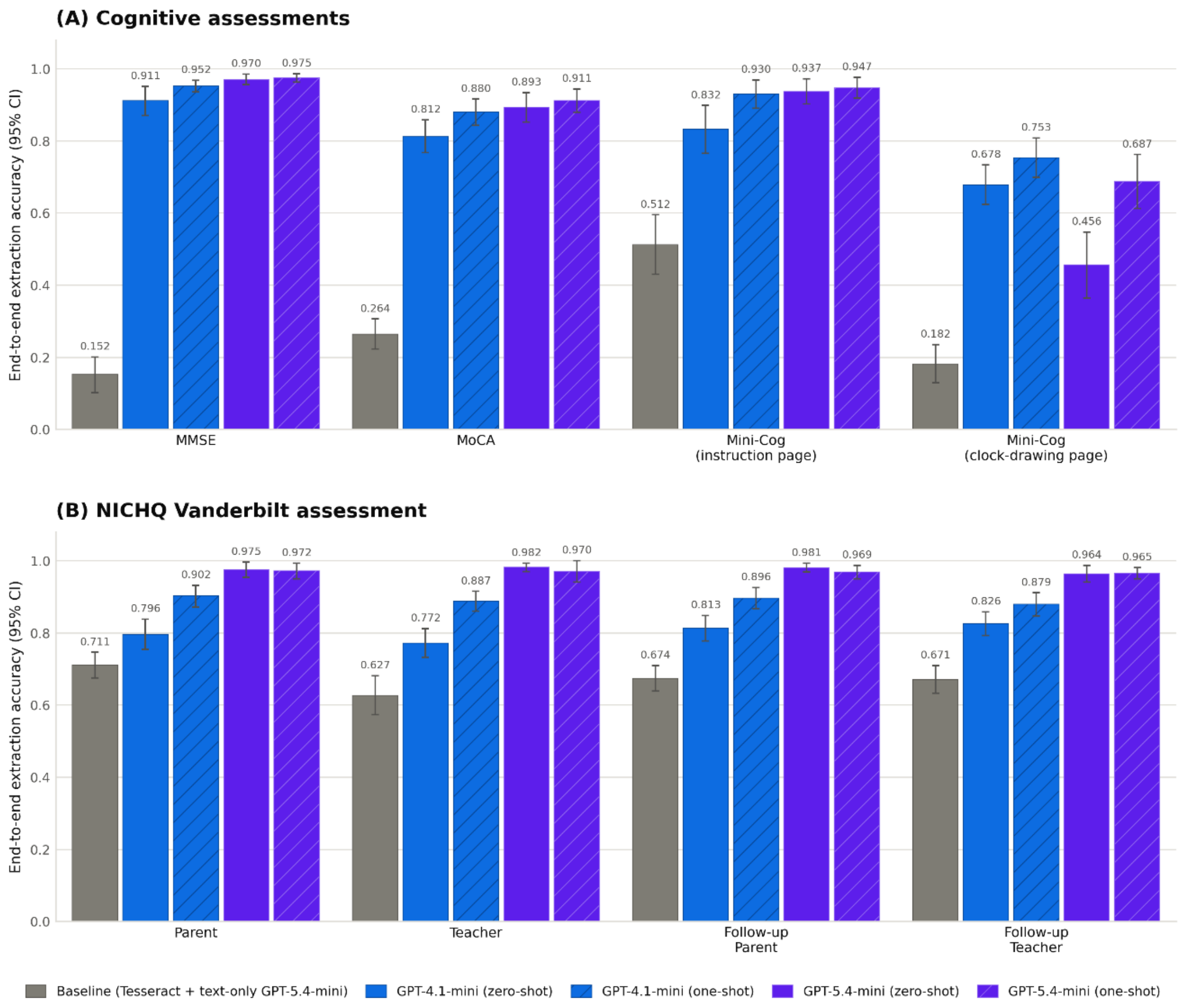

***Figure 4: End-to-end model performance. (A) Cognitive assessments**. GPT-5.4-mini with one-shot achieved the highest end-to-end score-extraction accuracy, dramatically outperforming the Tesseract baseline. **(B) NICHQ Vanderbilt Assessment**. GPT-5.4-mini with zero-shot achieved the highest end-to-end score-extraction accuracy.*

### Modular Evaluation of Form Classification and Score Extraction

For the cognitive assessments, overall, the GPT-4.1-mini with zero-shot prompting showed the best form-classification performance (macro F1=0.977). The GPT-5.4-mini one-shot showed a dominant score extraction (Acc=0.914, 95%CI=[0.898, 0.930]). Stratified by form types, GPT-4.1-mini zero-shot showed increased F1 scores, ranging from 0.977 to 1.000, for form classification in most types, while GPT-5.4-mini one-shot showed a consistent dominance in score extraction (Acc ranged from 0.779 to 0.976) (Table 2). For the NICHQ Vanderbilt Assessment, the GPT-5.4-mini zero-shot showed an overall dominance in both form classification (macro F1=0.970) and score extraction (Acc=0.978, 95%CI= [0.971, 0.984]). Stratified by assessment types, the model also achieved top performance for form classification (F1 range from 0.921 to 1.000) and score extraction (Acc range from 0.970 to 0.983) (Table 3).

*Table 2: Modular and end-to-end evaluation on cognitive assessments*

| Model | Modular evaluation | | End-to-end evaluation |
|---|---|---|---|
| **Overall** | | | |
| | **Form classification (Macro F1)** | **Score extraction, Acc [95%CI]** | **Score extraction, Acc [95%CI]** |
| Baseline | 0.840 | 0.334 [0.303, 0.366] | 0.278 [0.245, 0.310] |
| GPT-4.1-mini (zero-shot) | **0.977** | 0.810 [0.784, 0.836] | 0.808 [0.781, 0.835] |
| GPT-4.1-mini (one-shot) | 0.970 | 0.893 [0.876, 0.910] | 0.879 [0.858, 0.900] |
| GPT-5.4-mini (zero-shot) | 0.878 | 0.895 [0.877, 0.913] | 0.814 [0.780, 0.848] |
| GPT-5.4-mini (one-shot) | 0.941 | **0.914 [0.898, 0.930]** | **0.880 [0.856, 0.905]** |
| **MMSE** | | | |
| | **Form classification (F1)** | **Score extraction, Acc [95%CI]** | **Score extraction, Acc [95%CI]** |
| Baseline | **0.974** | 0.149 [0.100, 0.198] | 0.152 [0.102, 0.201] |
| GPT-4.1-mini (zero-shot) | 0.964 | 0.915 [0.875, 0.955] | 0.911 [0.870, 0.951] |
| GPT-4.1-mini (one-shot) | 0.969 | 0.951 [0.935, 0.967] | 0.952 [0.936, 0.968] |
| GPT-5.4-mini (zero-shot) | 0.964 | 0.969 [0.955, 0.984] | 0.970 [0.956, 0.985] |
| GPT-5.4-mini (one-shot) | 0.969 | **0.976 [0.965, 0.988]** | **0.975 [0.963, 0.986]** |
| **MoCA** | | | |
| | **Form classification (F1)** | **Score extraction, Acc [95%CI]** | **Score extraction, Acc [95%CI]** |
| Baseline | **0.984** | 0.266 [0.217, 0.314] | 0.264 [0.223, 0.306] |
| GPT-4.1-mini (zero-shot) | **0.984** | 0.825 [0.783, 0.866] | 0.812 [0.767, 0.858] |
| GPT-4.1-mini (one-shot) | 0.974 | 0.904 [0.880, 0.929] | 0.880 [0.843, 0.916] |
| GPT-5.4-mini (zero-shot) | 0.974 | 0.895 [0.862, 0.927] | 0.893 [0.852, 0.934] |
| GPT-5.4-mini (one-shot) | 0.974 | **0.921 [0.896, 0.946]** | **0.911 [0.879, 0.944]** |
| **Mini-Cog (Instruction page)** | | | |
| | **Form classification (F1)** | **Score extraction, Acc [95%CI]** | **Score extraction, Acc [95%CI]** |
| Baseline | **1.000** | 0.526 [0.443, 0.610] | 0.512 [0.430, 0.595] |
| GPT-4.1-mini (zero-shot) | **1.000** | 0.821 [0.758, 0.884] | 0.832 [0.765, 0.898] |
| GPT-4.1-mini (one-shot) | 0.979 | 0.940 [0.908, 0.973] | 0.930 [0.890, 0.969] |
| GPT-5.4-mini (zero-shot) | **1.000** | 0.937 [0.907, 0.967] | 0.937 [0.902, 0.972] |
| GPT-5.4-mini (one-shot) | **1.000** | **0.954 [0.925, 0.983]** | **0.947 [0.919, 0.976]** |
| **Mini-Cog (Clock-drawing page)** | | | |
| | **Form classification (F1)** | **Score extraction, Acc [95%CI]** | **Score extraction, Acc [95%CI]** |
| Baseline | 0.527 | 0.395 [0.358, 0.433] | 0.182 [0.129, 0.235] |
| GPT-4.1-mini (zero-shot) | **0.995** | 0.680 [0.626, 0.734] | 0.678 [0.624, 0.733] |
| GPT-4.1-mini (one-shot) | 0.984 | 0.777 [0.732, 0.823] | **0.753 [0.699, 0.808]** |
| GPT-5.4-mini (zero-shot) | 0.699 | **0.779 [0.734, 0.824]** | 0.456 [0.364, 0.547] |
| GPT-5.4-mini (one-shot) | 0.902 | 0.805 [0.762, 0.847] | 0.687 [0.612, 0.762] |
| **Other** | | | |
| | **Form classification (F1)** | **Score extraction, Acc [95%CI]** | **Score extraction, Acc [95%CI]** |
| Baseline | 0.716 | N/A | N/A |
| GPT-4.1-mini (zero-shot) | 0.940 | N/A | N/A |
| GPT-4.1-mini (one-shot) | **0.945** | N/A | N/A |
| GPT-5.4-mini (zero-shot) | 0.752 | N/A | N/A |
| GPT-5.4-mini (one-shot) | 0.860 | N/A | N/A |

*Table 3: Modular and end-to-end evaluation on NICHQ Vanderbilt Assessment*

| **Model** | **Modular evaluation** | | **End-to-end evaluation** |
|---|---|---|---|
| **Overall** | | | |
| | **Form classification (Macro F1)** | **Score extraction, Acc [95%CI]** | **Score extraction, Acc [95%CI]** |
| Baseline | 0.964 | 0.673 [0.652, 0.693] | 0.671 [0.650, 0.691] |
| GPT-4.1-mini (zero-shot) | 0.945 | 0.804 [0.786, 0.822] | 0.802 [0.783, 0.820] |
| GPT-4.1-mini (one-shot) | 0.947 | 0.894 [0.880, 0.907] | 0.891 [0.876, 0.906] |
| GPT-5.4-mini (zero-shot) | **0.970** | **0.978 [0.971, 0.984]** | **0.975 [0.967, 0.984]** |
| GPT-5.4-mini (one-shot) | 0.966 | 0.974 [0.966, 0.982] | 0.969 [0.958, 0.980] |
| **Parent** | | | |
| | **Form classification (F1)** | **Score extraction, Acc [95%CI]** | **Score extraction, Acc [95%CI]** |
| Baseline | **0.974** | 0.724 [0.695, 0.753] | 0.711 [0.675, 0.746] |
| GPT-4.1-mini (zero-shot) | 0.969 | 0.814 [0.784, 0.844] | 0.796 [0.754, 0.838] |
| GPT-4.1-mini (one-shot) | 0.959 | 0.913 [0.894, 0.932] | 0.902 [0.871, 0.932] |
| GPT-5.4-mini (zero-shot) | 0.954 | **0.983 [0.977, 0.989]** | **0.975 [0.954, 0.996]** |
| GPT-5.4-mini (one-shot) | 0.959 | **0.983 [0.978, 0.989]** | 0.972 [0.950, 0.993] |
| **Teacher** | | | |
| | **Form classification (F1)** | **Score extraction, Acc [95%CI]** | **Score extraction, Acc [95%CI]** |
| Baseline | 0.952 | 0.646 [0.599, 0.693] | 0.627 [0.573, 0.681] |
| GPT-4.1-mini (zero-shot) | 0.955 | 0.782 [0.743, 0.821] | 0.772 [0.732, 0.812] |
| GPT-4.1-mini (one-shot) | 0.960 | 0.889 [0.861, 0.916] | 0.887 [0.860, 0.915] |
| GPT-5.4-mini (zero-shot) | **0.979** | **0.979 [0.964, 0.995]** | **0.982 [0.970, 0.993]** |
| GPT-5.4-mini (one-shot) | 0.969 | 0.977 [0.959, 0.994] | 0.970 [0.941, 1.000] |
| **Follow-up Parent** | | | |
| | **Form classification (F1)** | **Score extraction, Acc [95%CI]** | **Score extraction, Acc [95%CI]** |
| Baseline | 0.984 | 0.664 [0.627, 0.701] | 0.674 [0.639, 0.709] |
| GPT-4.1-mini (zero-shot) | 0.964 | 0.809 [0.771, 0.848] | 0.813 [0.778, 0.848] |
| GPT-4.1-mini (one-shot) | 0.974 | 0.885 [0.850, 0.919] | 0.896 [0.867, 0.926] |
| GPT-5.4-mini (zero-shot) | **1.000** | **0.978 [0.965, 0.990]** | **0.981 [0.969, 0.993]** |
| GPT-5.4-mini (one-shot) | 0.995 | 0.973 [0.959, 0.987] | 0.969 [0.950, 0.987] |
| **Follow-up Teacher** | | | |
| | **Form classification (F1)** | **Score extraction, Acc [95%CI]** | **Score extraction, Acc [95%CI]** |
| Baseline | 0.979 | 0.656 [0.610, 0.702] | 0.671 [0.633, 0.709] |
| GPT-4.1-mini (zero-shot) | 0.979 | 0.812 [0.775, 0.849] | 0.826 [0.793, 0.858] |
| GPT-4.1-mini (one-shot) | 0.979 | 0.888 [0.861, 0.914] | 0.879 [0.847, 0.911] |
| GPT-5.4-mini (zero-shot) | **0.995** | **0.970 [0.952, 0.989]** | **0.964 [0.941, 0.987]** |
| GPT-5.4-mini (one-shot) | 0.984 | 0.964 [0.942, 0.987] | 0.965 [0.950, 0.981] |
| **Other** | | | |
| | **Form classification (F1)** | **Score extraction, Acc [95%CI]** | **Score extraction, Acc [95%CI]** |
| Baseline | **0.930** | N/A | N/A |
| GPT-4.1-mini (zero-shot) | 0.857 | N/A | N/A |
| GPT-4.1-mini (one-shot) | 0.864 | N/A | N/A |
| GPT-5.4-mini (zero-shot) | 0.921 | N/A | N/A |
| GPT-5.4-mini (one-shot) | 0.921 | N/A | N/A |

**NICHQ Vanderbilt Assessment Text Fields Extraction**

We prompted LLMs to extract the free-text comments (across all four form types) and the class name (on the Teacher and Follow-up Teacher forms). For comment extraction, the GPT-5.4-mini zero-shot achieved an overall recall of 0.970, precision of 0.737, and string similarity to the ground truth of 0.770. For class name extraction, an overall 0.988 sensitivity, 0.871 precision, and 0.888 text similarity were achieved (Tables S11 and S12).

**Error Analysis**

Each erroneous field of the GPT-5.4-mini one-shot outputs is reviewed and assigned a cause based on our judgment. For the cognitive assessments, the main cause of the 260 errors is misreading a number due to the LLM's visual capacity limits (N=90, 34.6%). **Ambiguous drawings** on the Mini-Cog clock-drawing page contributed to 31.2% (N=81) of errors. We performed an in-depth analysis in the following section. **Hallucinations** in which the LLM generates non-existent numbers are another common cause, accounting for 13.8% (N=36) of errors. **Overwritten or crossed-out numbers**, **illegible handwriting**, and **poor scanning quality** contributed 7.7% (N=20), 5.0% (N=13), and 3.1% (N=8), respectively. For the 497 errors in the NICHQ Vanderbilt Assessment, **poor scanning quality**, such as rotation, blurriness, and background noise, caused 23.9% (N=119) errors. LLM **misreading marks** caused 19.5% (N=97) errors. **Hallucination** caused 17.3% (N=86). **Overwritten or corrected marks**, **unclear marks**, and **LLM reading the wrong rows** accounted for 8.2% (N=41), 5.4% (N=27), and 4.6% (N=23) of all errors, respectively. Examples of each error type are presented in Figure 5.

| Types of Error | Examples | Ground truth | Model |
|---|---|---|---|
| **(A)** Rotated scan | 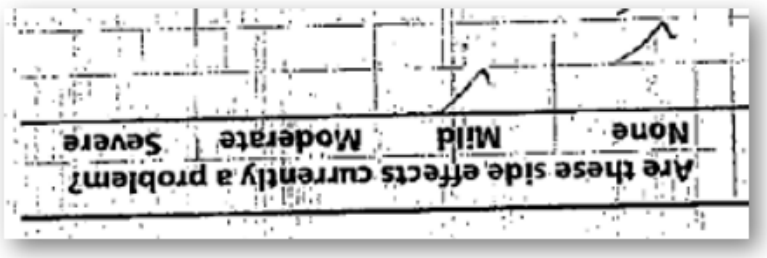 | Mild | [Blank] |
| **(B)** Faint scan | 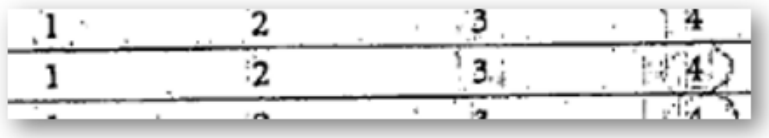 | 4 | 3 |
| **(C)** Overwritten mark | 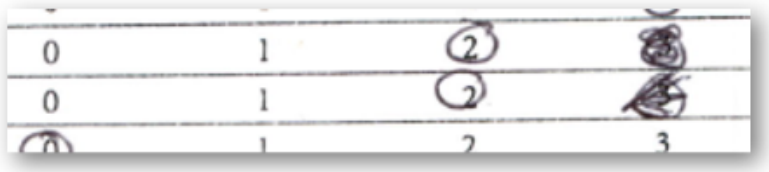 | 2 | 3 |
| **(D)** Overwritten number | 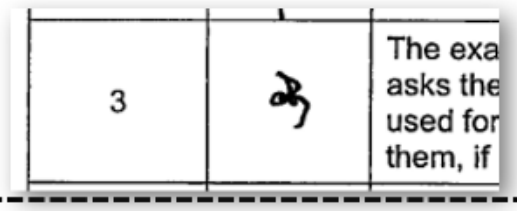 | 3 | 2 |
| **(E)** Model misread mark | 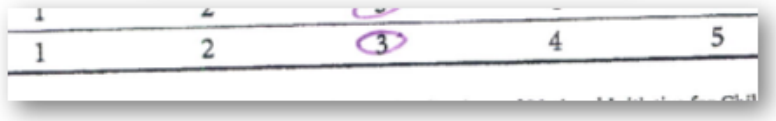 | 3 | 4 |
| **(F)** Model misread number | 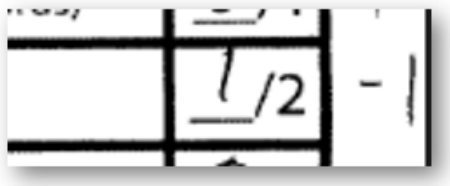 | 1 | 2 |
| **(G)** Hallucination | 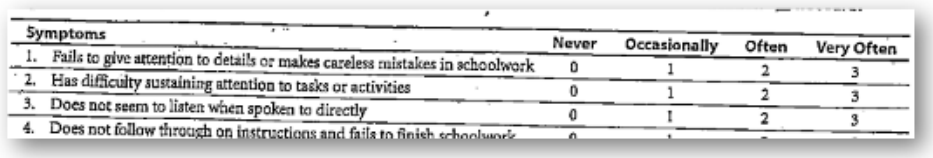 | [Blank]<br>[Blank]<br>[Blank]<br>… | 1<br>2<br>0<br>… |
| **(H)** Illegible handwriting | 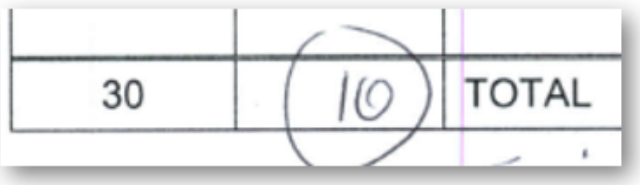 | 16 | 10 |
| **(I)** Ambiguous marks | 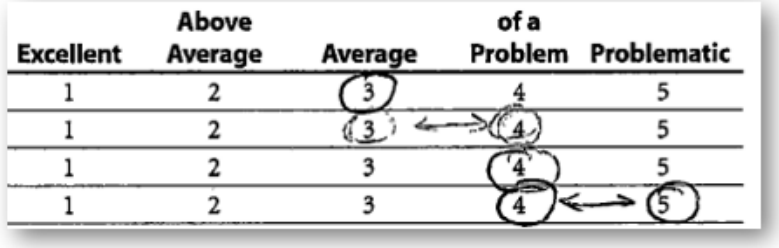 | 3<br>4<br>4<br>5<br>… | 3<br>[Blank]<br>[Blank]<br>[Blank]<br>… |
| **(J)** Model read wrong row/field | 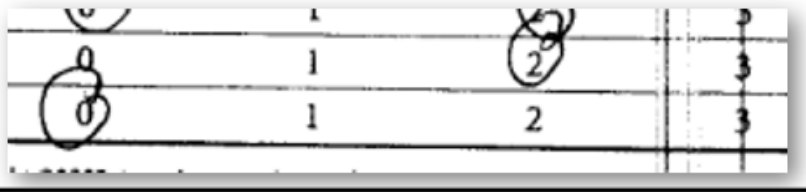 | 2<br>0 | 0<br>0 |

***Figure 5: Types of errors**. Incorrect score extraction is marked in red: **(A) Rotated scan**, which, with an 180-degree rotation during scanning, causes a systematic decrease in extraction performance. **(B) Faint scan** with faint text and background noise. **(C) Overwritten mark,** which overwritten or corrected marks cause misreading. **(D) Overwritten numbers**, which are overwritten or corrected, cause misreading. **(E) Model misreads marks**, which the LLM misreads marks due to its vision capacity limits. **(F) Model misreads numbers**. LLM misreads numbers or text due to its visual processing limitations. **(G) Hallucination**. LLM hallucinates marks that do not exist. **(H) Illegible handwriting**. Messy handwriting causes extraction failure. **(I) Ambiguous marks**. Unclear marks or marks including multiple scores. **(J) Model read the wrong row/fields**. LLM extracted scores from a neighboring row.*

### Qualitative Analysis of Mini-Cog Clock-drawing Ambiguous Cases

Among the 130 erroneous field and score extractions on the Mini-Cog clock-drawing pages, 62.3% (N=81) were related to ambiguous marks and drawings. The most common source of ambiguity is the **indistinguishable long and short arms**, which account for 30.8% (N=40) of all errors. The two arms were drawn with similar lengths, making the time on the clock unreadable. For example, with one arm pointing at “12” and the other at “4”, without knowing which arm is longer, it can be read as “12:20” or “4:00”. The second most common problem is that **arms point at ambiguous numbers**, which accounts for 12.3% (N=16). A short arm pointing between numbers “10” and “11” and the long arm pointing at “3” can be read as “10:15” or “11:15”. **Misaligned angles and numbers** accounts for 11.5% (N=15) of errors. The numbers are drawn in the wrong position. Reading the numbers and the angles of the arms yields different results. For example, when the number “4” is placed at the bottom of the clock and the number “8” is placed on the top-left, reading the numbers results in "8:20” while reading the arm angles results in “10:30”. Finally, some clock drawings include **multiple long and short arms**, which account for 7.7% (N=10) of errors. For example, two short arms are drawn pointing to the numbers “10” and “11”, leaving the time on the clock undetermined. Overall, in those ambiguous cases, despite deviations from the human-annotated ground truth, the LLM demonstrated a comprehensive understanding of the drawing and produced reasonable extractions (Figure 6).

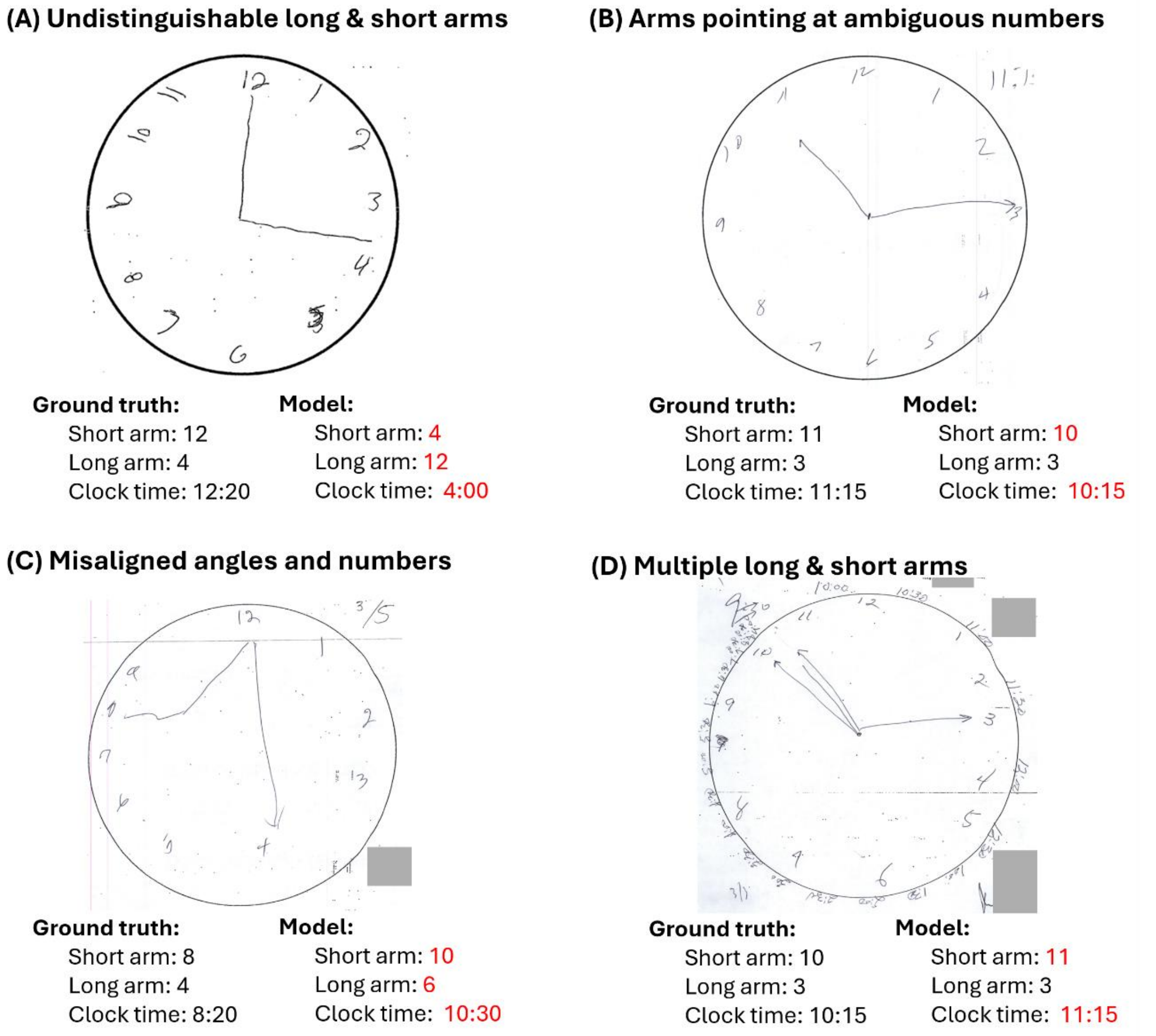

***Figure 6: Ambiguous cases of Mini-Cog clock-drawing**. Affected LLM outputs are marked in red. **(A) Indistinguishable long and short arms**. The long and short arms are similar in length, making the time on the clock ambiguous. **(B) Arms pointing at ambiguous numbers**. The arms point between numbers, leaving the time on the clock undetermined. **(C) Misaligned angles and numbers**. The numbers are not written in the regular positions. **(D) Multiple long and short arms** are drawn on the clock.*

### Evaluation of the Human-in-the-loop Design

Using GPT-4.1-mini and GPT-5.4-mini with one-shot prompting, the dual-model validation achieved an overall coverage of 0.638 across all five cognitive form types, including Other. Among pages where both models agreed on form type and extracted scores, the mean extraction accuracy was 0.975. Compared with the best single model, a human-in-the-loop design resulted in a 9.5% overall improvement. For the MMSE, MoCA, and Mini-Cog, this design showed coverage of 0.368 to 0.821 while boosting end-to-end accuracy to 0.938 to 0.987 (Table 5).

*Table 4: Evaluation of the human-in-the-loop design for cognitive assessments*

| Forms | GPT-4.1-mini (one-shot) + GPT-5.4-mini (one-shot) | | Best single model |
|---|---|---|---|
| | **Proportion of pages agreed by both models, Coverage [95%CI]** | **End-to-end form extraction on agreed pages, Acc [95%CI]** | **End-to-end form extraction, Acc [95%CI]** |
| Overall | 0.638 [0.594, 0.680] | 0.975 [0.964, 0.986] | 0.880 [0.856, 0.905] |
| MMSE | **0.695 [0.596, 0.778]** | 0.985 [0.973, 0.997] | 0.975 [0.963, 0.986] |
| MoCA | 0.368 [0.278, 0.469] | 0.969 [0.952, 0.986] | 0.911 [0.879, 0.944] |
| Mini-Cog (Instruction page) | 0.821 [0.732, 0.885] | 0.987 [0.973, 1.002] | 0.947 [0.919, 0.976] |
| Mini-Cog (Clock-drawing page) | 0.389 [0.298, 0.490] | 0.938 [0.889, 0.988] | 0.753 [0.699, 0.808] |
| Other | 0.916 [0.843, 0.957] | NA | NA |

**Cost Analysis**

This section summarizes processing time and cost. A more comprehensive analysis is reported in the Supplementary *Implementation Details — Cost Analysis* section.

**Cognitive assessments**. We estimate it would take about 74 hours for human annotators to label a pool of 10,000 cognitive assessment forms, assuming equal distribution of types and assuming no “other” types. Using a wage estimate of $20.05 per hour, we estimate the cost to be $1,480. Using the GPT-5.4-mini API rate on Microsoft Azure, the estimated cost for the one-shot approach is $122 (Table S17) with 6.1 hours to process. The Human-in-the-loop approach requires 12.2 hours of machine time (dual-model agreement step) and 26.7 hours of manual time (human verification step), and costs $722.

**NICHQ Vanderbilt assessment**. We estimate it would take hours for human annotators to label a pool of 10,000 forms, assuming equal distribution types, costing approximately $2921. GPT-5.4-mini with one-shot prompting takes about 7.6 hours while costing about $155 (Table S18).

## DISCUSSION

This study proposed a visual information extraction system for paper-based form processing using multimodal LLMs and achieved over 90% accuracy for most cognitive assessments and over 96% accuracy for NICHQ Vanderbilt Assessment versions, dramatically outperforming the traditional OCR engine. Compared to the current manual data extraction practice, our system reduces processing time by 10 to 20 times and costs by 12 to 18 times. Our human-in-the-loop design further demonstrates a practical solution that reduces manual work by two-thirds while boosting end-to-end accuracy to over 97.5%. Our modular design maximizes flexibility in continuous system development and maintenance. To add a new form type, one would simply amend the classifier prompt and add a new score extractor, while the rest of the system remains unchanged. While this system was evaluated using cognitive and ADHD assessments, it is generalizable to other paper-based forms with an appropriate prompt design.

A recent study using multimodal LLMs to process transfusion reaction reports has shown an overall accuracy of 92.0% and an accuracy of 87.5% on faint, corrected, or ambiguous handwritten markings [6]. Our previous multimodal LLM study on scanned death certificates showed an 86.3% overall accuracy for the extraction of the immediate cause of death and an 75.7% accuracy for handwritten ones [7]. Building on prior work of single-typed paper-based forms, this study further developed a modular pipeline and a human-in-the-loop solution to enhance generalizability and performance.

The lack of structured data extraction is a long-standing challenge for healthcare organizations. At our organization, only 0.5% of the 1,204 patients aged 50 years and older who had a scanned Mini-Cog in 2025 had the data captured in structured fields. Similarly, among the 1,064 pediatric patients who received care between May 2021 and October 2024 and had a documented ADHD screening, only 262 (24.6%) had results documented as structured data. Our proposed system provides a promising solution for an accurate, scalable, and cost-efficient structured data extraction.

The error analysis reveals that, despite largely improved performance, LLMs' visual capacity remains a challenge for some visual information extraction tasks and accounts for 34.6% and 19.5% of all errors on cognitive assessments and the NICHQ Vanderbilt assessment, respectively. Hallucination is the second most frequent factor, accounting for 13.8% and 17.3%, respectively. Interestingly, our qualitative analysis of the Mini-Cog Clock Drawing revealed the LLM's surprising spatial reasoning capacity. Among 130 erroneous field extractions on Mini-Cog clock-drawing pages, 81 (62.3%) were associated with ambiguous drawings. Some model outputs were plausible alternatives to the reference annotations. These findings underscore uncertainty in assigning a single label to visually ambiguous features. Explicit ambiguity rules and expert adjudication should be considered. This preliminary observation provides a starting point for future studies on non-textual visual information extraction and AI-driven diagnostics.

Despite the promising results, this study has a few limitations. First, only two proprietary LLMs, GPT-4.1-mini and GPT-5.4-mini, were evaluated. At this point, the only Health Insurance Portability and Accountability Act (HIPAA)-compliant LLM service available to us is Microsoft Azure OpenAI. Future studies are needed to explore other multimodal LLMs, including open-weight models. Second, different versions of the cognitive and Vanderbilt assessments contain slightly different wording and layouts. For simplicity, we used a static version of each form for system development. In our datasets, the differences among different versions are negligible. However, a careful comparison and data checking are needed before adoption. Third, reasoning tokens from GPT-5.4-mini are not available to end users; therefore, our cost estimate slightly understates the actual cost. However, since the reasoning effort was set to a low level, the difference should be minimal. Finally, as a single-organization study, generalizability may be a challenge. However, as a large health system with over 279 clinic staff who create scanned documents on a daily basis, we believe this issue is mitigated by the diverse range of writing styles and scanning artifacts.

## CONCLUSIONS

Multimodal LLMs can unlock clinically valuable data currently trapped in scanned handwritten paper forms. Our LLM-based modular system achieved high extraction accuracy across cognitive and behavioral assessment forms, outperforming traditional OCR while reducing processing time and cost by more than an order of magnitude. Our human-in-the-loop design, by accepting concordant outputs and directing only discordant or ambiguous cases to human review, provides a scalable and practical framework for converting unstructured, handwritten data into actionable data for clinical decision support. Our Python package, production code, and prompts are publicly available.

## ACKNOWLEDGEMENTS

The research reported in this manuscript was supported by AIM-AHEAD Coordinating Center at the University of North Texas Health Science Center at Fort Worth. This research was, in part, funded by the National Institutes of Health (NIH) Agreement No. 1OT2OD032581 (Principal Investigators: Yen-Chi Le and Sandra Stansberry) and NIH Grant No. R01LM014508 (Principal Investigator: Kirk Roberts). The views and conclusions contained in this document are those of the authors and should not be interpreted as representing the official policies, either expressed or implied, of the NIH.

The authors would like to thank our clinical code signers and subject matter experts: Julie Bortolotti, Kevin Hwang, Thomas J. Murphy, Eric Thomas, etc. We thank our annotators: Yunbo Xie, Paris Nix, and Mary Newton.

## COMPETING INTERESTS

The authors have no competing interests to disclose.

## CONTRIBUTORSHIP

**Enshuo Hsu:** Conceptualization, Methodology, Software, Visualization, Formal analysis, Data curation, Writing – original draft, Writing – review & editing. **Linh Nguyen:** Writing – review & editing, Methodology. **Sunyang Fu:** Methodology, Writing – review & editing. **Heath Goodrum:** Methodology, Data curation, Writing – review & editing. **Sandra Stansberry:** Funding acquisition, Methodology, Writing – review & editing. **Kirk Roberts:** Supervision, Writing – review & editing. **Hongfang Liu:** Writing – review & editing. **Yen-Chi L. Le:** Conceptualization, Project administration, Funding acquisition, Supervision, Writing – review & editing.

## DATA AVAILABILITY STATEMENT

The cognitive and ADHD assessment scans contain PHI. Releasing is prohibited by our institutional policy. For reproducibility, we synthesized a set of 50 scanned MMSE, MoCA, Mini-Cog, and NICHQ Vanderbilt forms that do not contain real patient information and shared them on our GitHub repository (https://github.com/daviden1013/Cognitive_ADHD_VLM/) along with our experimental code.

## FUNDING

This research was, in part, funded by the National Institutes of Health (NIH) Agreement No. 1OT2OD032581 (Principal Investigators: Yen-Chi Le and Sandra

Stansberry) and NIH Grant No. R01LM014508 (Principal Investigator: Kirk Roberts).